%% file: paper.tex
\documentclass{article}
\usepackage{ijcai22}
\usepackage{times}
\usepackage{soul}
\usepackage{url}
\usepackage[hidelinks]{hyperref}
\usepackage[utf8]{inputenc}
\usepackage[small]{caption}
\usepackage{graphicx}
\usepackage{amsmath}
\usepackage{amssymb}
\usepackage{amsthm}
\usepackage{booktabs}
\usepackage{algorithmic}
\usepackage[ruled,linesnumbered]{algorithm2e}
\usepackage{xcolor}
\makeatletter
\renewcommand{\@algocf@capt@plain}{above}
\makeatother

\newcommand{\bE}{\mathbb{E}}

\newcommand{\BR}{\textsc{BR}}

\newcommand{\cA}{\mathcal{A}}

\newcommand{\cN}{\mathcal{N}}

\newcommand{\cS}{\mathcal{S}}
\newcommand{\cT}{\mathcal{T}}
\newcommand{\cZ}{\mathcal{Z}}

\newcommand{\defword}[1]{\textbf{\boldmath{#1}}}

\newcommand{\ABR}{\textsc{ABR}}
\makeatletter
\newcommand\footnoteref[1]{\protected@xdef\@thefnmark{\ref{#1}}\@footnotemark}
\makeatother

\definecolor{darkgreen}{RGB}{0,125,0}

\newcounter{mlNoteCounter}

\newcounter{rmNoteCounter}

\newcounter{ftNoteCounter}

\newcounter{neilNoteCounter}

\newcounter{nolanNoteCounter}

\title{Approximate Exploitability: Learning a Best Response}

\author{
    Finbarr Timbers$^1$\and
    Nolan Bard$^1$ \and
    Edward Lockhart$^1$\and
    Marc Lanctot$^1$\and
    Martin Schmid$^1$\footnote{Work was done while at DeepMind. Currently at Equilibre Technologies.}\and
    Neil Burch$^{1, 2}$\and
    Julian Schrittweiser$^1$\and
    Thomas Hubert$^1$\And
    Michael Bowling$^{1,2}$\\
    \affiliations
    $^1$DeepMind\\
    $^2$University of Alberta\\
    \emails
    finbarrtimbers@google.com
}

\begin{document}

\maketitle

\begin{abstract}
Researchers have shown that neural networks are vulnerable to adversarial examples and subtle environment changes.  The resulting errors can look like blunders to humans, eroding trust in these agents. In prior games research, agent evaluation often focused on the in-practice game outcomes. Such evaluation typically fails to evaluate robustness to worst-case outcomes. Computer poker research has examined how to assess such worst-case performance. Unfortunately, exact computation is infeasible with larger domains, and existing approximations are poker-specific. We introduce ISMCTS-BR, a scalable search-based deep reinforcement learning algorithm for learning a best response to an agent,  approximating worst-case performance. We demonstrate the technique in several games against a variety of agents, including several AlphaZero-based agents. Supplementary material is available at https://arxiv.org/abs/2004.09677.
\end{abstract}

\section{Introduction}

The increasing incorporation of artificial intelligence into the systems of human life has created a growing impetus to understand the safety and robustness implications of using these algorithms in critical systems.  Unexpected failures of AI systems will erode trust, particularly in high consequence settings like medicine and autonomous vehicles.  
Researchers have been turning their attention to this problem~\cite{AmodeiOSCSM16Safety}, highlighting how neural networks trained by deep learning techniques can suffer from surprising failure modes due to distribution shift~\cite{Quionero-Candela09}: the situation where a network is trained based on one distribution of data, but then used under a different distribution. This phenomenon has presented itself in a variety of areas, including adversarial examples in image classification~\cite{AthalyeEIK18}, and reinforcement learning in both single agent~\cite{WittyLTALJ18} and multi-agent settings~\cite{GleaveDWKLR20}.  

Artificial intelligence research has long used games as benchmarks for progress, including milestone achievements in Checkers \cite{schaeffer1996chinook}, Chess \cite{campbell2002deep}, Poker \cite{Moravcik17DeepStack,Brown17Libratus}, Go \cite{Silver18AlphaZero}, and Atari \cite{mnih2015human}, among others. In these games, agent evaluation is commonly based on the outcomes generated in practice through play.  In single agent domains, a game score such as in Atari~\cite{bellemare13arcade} is often used to assess quality.  However, in multi-agent domains, the in-practice performance of an agent depends directly on the assumed population of other agents~\cite{bard2016online}.  Furthermore, agent performance is often not strictly transitive~\cite{Tuyls18Generalized} – with agent A beating B, B beating C, and C beating A – making measures such as Elo problematic~\cite{Balduzzi18,Omidshafiei19}. 

While such evaluation is important, less attention is typically given to the potential worst-case performance of an agent.  Dominant algorithms for two-player zero-sum games rely on self-play to train an agent, resulting in agents that have observed a particular distribution of play.  However, other agents may induce distribution shift through entirely valid play, either deliberately due to the adversarial nature of these domains, or simply through sub-optimal play.  This makes an agent’s robustness to worst-case performance an important complementary dimension of agent evaluation.  Furthermore, in light of the low consequence, repeatable, and controllable nature of games, we believe exploring agent robustness and AI safety concerns in these domains is another opportunity for games research to drive AI forward as safety considerations grow in importance.

Prior research in computer poker has attempted to directly evaluate an agent’s performance against a worst-case (i.e. best responding) adversary, as game-theoretic algorithms approximating a Nash equilibrium are standard~\cite{08nips-cfr} and seek to optimize this (for two-player zero-sum games). Exactly computing a policy’s \defword{exploitability} – its loss relative to a Nash in the worst-case – was feasible in heads-up limit Texas hold’em poker (HUL)~\cite{johanson2011accelerating}, having roughly $10^{14}$ decision points.  Unfortunately, other common games are vastly larger, such as heads-up no-limit Texas hold’em (HUNL) (with around $10^{160}$ decision points~\cite{johanson2013measuring}) or Go (approximately $10^{170}$ decision points~\cite{Allis94thesis}). In these games, computing exploitability or the best response strategy is intractable.  However, the \defword{local best response} (LBR)~\cite{Lisy16LBR} approximation was used to provide a lower-bound on the exploitability of HUNL strategies.
LBR uses explicit knowledge of the distribution over its opponent’s private state with a shallow search and simple poker-specific heuristic value function.  Though simple, LBR’s reliance on domain knowledge and notable sensitivity to the various choices of abstractions and heuristics make it challenging to use broadly.

We generalize the approach introduced by LBR, introducing a scalable domain-independent alternative that uses deep reinforcement learning (RL) combined with search and knowledge of the opponent’s distribution over private states to learn a value function for the best response in a two-player zero-sum game.  By fixing the policy of the agent being evaluated, the environment becomes single-agent from the perspective of the agent, and the optimal policy from RL is exactly a best response \cite{greenwald2017solving}. By exploiting access to the agent’s policy during evaluation, we hope to simplify the learning task, reducing the representational power needed by the network, and improving sample efficiency and accuracy.

Our contributions are as follows:

\begin{itemize}
    \item In Section~\ref{sec:ISMCTS-BR}, we introduce our best response algorithm, \defword{ISMCTS-BR}.
    \item In Section~\ref{sec:validation}, we demonstrate the robustness of ISMCTS-BR across a large number of games \& policies by showing that it is able to closely approximate the exact best response.
    \item In Section~\ref{sec:AZ-go}, we demonstrate that ISMCTS-BR can exploit strong, search-based agents, and provide evidence that deep reinforcement learning techniques can be vulnerable to distribution shift, highlighting the importance of worst-case performance measures in agent evaluation.
\end{itemize}

\section{Background and Terminology}
\label{sec:background}
An \defword{extensive-form game} is a sequential interaction between \defword{players} $i \in \{ 1, 2, \cdots, n\} \cup \{ c \}$, where $c$ is the \defword{chance player}, i.e. a player following a static, stochastic policy to define the transition probabilities given states and actions. We use $-i$ to refer to the non-chance opponent of player $i$. In this paper, we focus on the two-player zero-sum setting, and in particular, those where beliefs about the opponent's private state are tractable, i.e. sufficiently small to be represented in memory.
The game starts in the root (or empty) history $h = \emptyset$. On each turn, a player $i$ (possibly chance) chooses an action $a \in \cA_i$, changing the history to $h' = ha$. 
The full history is sometimes also called a {\it ground} state because it uniquely identifies the true state, as it consists of all actions including those not observed by all players.
For example, a history in poker includes all the dealt cards, even private ones, in addition to the sequence of actions taken.
We define an \defword{information state} (or \textbf{infostate}) $s \in \cS$ for player $i$ as the state as perceived by an agent which is consistent with its observations.\footnote{An \textbf{information state} is equivalent to the concept of an information set within the literature on extensive form games.  We choose to use the phrase information state in order to emphasize its role as the object an agent’s policy conditions on when choosing an action. We find this more intuitive when used in a RL setting.}
Formally, each $s$ is a set of histories, specifically $h, h' \in s \Leftrightarrow$ the sequence of player $i$'s observations along $h$ and $h'$ are equal. 
Informally, this consists of all the histories which the player cannot distinguish given the information available to them.
We call $\cZ$ the set of terminal histories, which correspond to the end of a game. Each $z \in \cZ$ has a utility for each player $u_i(z)$.
Since players cannot observe the ground state $h$, policies are defined as $\pi_i : \cS_i \rightarrow \Delta(\cA_i)(s)$,
where $\Delta(\cA_i)(s)$ is the set of probability distributions over the legal actions available to player $i$ at $s$, $\cA_i(s)$.

For example, in Texas hold'em poker, the root history is when no cards have been dealt and no betting has happened. The history includes any private cards that have been dealt, while the information state for player $i$ includes the private cards for that player, and the information that the other player has received cards, but not which cards those are. Both the history and the information state include all public cards and all betting actions. A terminal history is a history where either a showdown has occurred or a player has folded. $u_i(z)$ is then the number of chips that each player has won. 

In a perfect information game such as Chess or Go, the history is identical to the information state, since everything is public.
As is common in perfect information games, we use an encoding of the current board state (or a stack of recent ones) as input for training, following \cite{Silver18AlphaZero}. $u_i(z)$ is -1 for a loss, 1 for a win, and 0 for a draw when possible (e.g. in Chess).
We assume finite games, so every history $h$ is bounded in length. 
The expected value of a joint policy $\pi$ (all players' policies) for player $i$ is defined as $u_{i}(\pi) = \bE_{z \sim \pi}[ u_i(z) ]$, 
where the terminal histories $z \in \cZ$ are composed of actions from the joint policy.

\subsection{Optimal Policies}
The standard game-theoretic solution concept for optimal policies in two-player zero-sum games is \textbf{Nash equlibrium}, defined as a strategy profile (i.e., joint policy) where none of the players benefits from deviating unilaterally. A strategy profile $\pi = (\pi_1, \pi_2)$ is a Nash equilibrium iff
\begin{align}
    \forall \pi'_i, i \in \{1,2\} : \; u_i(\pi_i, \pi_{-i}) \geq u_i(\pi'_i, \pi_{-i})
\end{align}
In two-player zero-sum games, this concept is equivalent to the $\textbf{minmax}$ solution concept, and the expected value for playing optimally is unique: 
\begin{align}
    v_1^* = \max_{\pi_1} \min_{\pi_2} u_1(\pi_1, \pi_2) = \min_{\pi_2} \max_{\pi_1} u_1(\pi_1, \pi_2).
\label{eq:minmax}    
\end{align}

The value (\ref{eq:minmax}) is referred to as the \textbf{game value} (for player 1), $v_1^*$. In two-player zero-sum games, the game value is thus $v_2^* = -v_1^*$.
Both solutions concepts thus optimize against the worst-case opponent --- the \defword{best response}.
We define the set of player $i$ best responses to an opponent policy $\pi_{-i}$,
\[
\BR_i(\pi_{-i}) = \{ \pi_i~|~ u_i(\pi_i, \pi_{-i}) = \max_{\pi_i'} u_i(\pi_i', \pi_{-i}) \}.
\]
For convenience, we refer to elements in this set, best responses to $\pi_{-i}$, as $b(\pi_{-i})$.
When a player uses an optimal policy $\pi^*_i$, their utility is then guaranteed (on expectation) to
achieve at least the game values: $\forall i, \pi_{-i}:  u_i(\pi^*_i, \pi_{-i}) \geq v^*_i$.

\subsection{Exploitability}

For suboptimal policies $\pi$, player $i$'s incentive to deviate is: 
\[
\delta_i(\pi) = u_i(b(\pi_{-i}), \pi_{-i}) - u_i(\pi).
\]
One metric to quantify the distance from optimal is \textsc{NashConv}$(\pi) = \sum_i \delta_i(\pi)$ and \textsc{Exploitability}$(\pi) =  \textsc{NashConv}(\pi)/n$.
In zero-sum games, when summing over players, the second terms sum to zero, so \textsc{NashConv} simplifies to $\sum_i u_i(b(\pi_{-i}), \pi_{-i})$.
Furthermore, the $\epsilon$-minmax (or $\epsilon$-Nash equilibrium) policy is one where $\max_i \delta_i(\pi) \le \epsilon$.




Exploitability measures how well a strategy profile approximates a Nash equilibrium --- the closer it is to zero, the closer the policy is to optimal.
Unfortunately, exploitability can be difficult to compute exactly: in the worst case, it requires traversing the full game tree, making it intractable in large games.  However, by approximating exploitability, we can address tractability while still assessing worst-case performance.

\section{Approximate Best Response}

By fixing the policy of one agent, the environment becomes a (stochastic) single agent environment
where one can use standard reinforcement learning methods to learn a best response, as they are optimal policies in that environment~\cite{greenwald2017solving}. The fixed agent is \emph{exploitable} by the reward obtained by the optimal policy.

While exploitability is defined using an exact best-responding policy $b(\pi_{-i})$, we can approximate this policy and define a corresponding metric, the approximate best response ($\ABR_i(\pi_{-i})$), with the corresponding $\textsc{ApproximateNashConv}$ ($\textsc{ANC}$) as:
\begin{equation}
\label{eq:anc}
\textsc{ANC}(\pi) = \sum_i u_i(\ABR_i(\pi_{-i}), \pi_{-i})
\end{equation}
We define approximate exploitability similarly. Since it uses an approximate best response, the approximate exploitability is a lower bound on the true value, with equality when the approximation matches the exact best response.

\subsection{Local Best Response}

The poker community has used local best response (LBR) \cite{Lisy16LBR} as a proxy for the full best response \cite{Moravcik17DeepStack,brown2018depth}.
LBR approximates the best response by performing a local search aided by explicit knowledge of the distribution over its opponent’s private state and a value function, typically a hand-crafted heuristic function.
LBR is fast to run and easy to understand; it can often be implemented in a small amount of code.



However, the results LBR obtains can be sensitive to the specific experimental setup. For example, when running against the same fixed policy using the same heuristic value approximator in action-abstracted no-limit poker, the values can vary wildly depending on the choice of action abstraction (e.g. from -867 mbb/h to 46 mbb/h~\cite{armac} and from 496 mbb/h to 3763 mbb/h~\cite{Lisy16LBR}). Additionally, while the check-call value function has had much empirical success in poker, it is not obvious what a strong choice for value function is in other games.

\subsection{IS-MCTS Best Response Search}
\label{sec:ISMCTS-BR}

LBR performs a depth-limited tree search, using a value function to truncate the search. This is similar to MCTS, which has been shown to perform remarkably well with a learned value function \cite{Silver18AlphaZero}. A natural extension to LBR then is to combine it with MCTS. Since we want an algorithm that works on imperfect information games, we use a variant of MCTS, information set MCTS (IS-MCTS)~\cite{Cowling12ISMCTS}. However, IS-MCTS does not sample from the belief distribution when searching at an information state. While empirically this works to exploit basic agents, this does not converge to the exact best response in the limit~\cite{Lisy15Online}.

Instead, we calculate the exact posterior distribution over histories with the same information state given the opponent's policy, and use that to sample from opponent information states.\footnote{In perfect information games, this is equivalent to MCTS.} To the best of our knowledge, ours is the first algorithm to apply IS-MCTS with the exact posterior distribution over opponent private states and PUCB. 

\begin{algorithm}[tb]
\caption{IS-MCTS best response search.}\label{alg:ismctsbr}
\SetKwInOut{Input}{input}\SetKwInOut{Output}{output}
\Input{$n$ --- number of simulations}
\Input{$\pi_{-i}$ --- fixed policy for opponent}
\Input{$s$ --- infostate to choose an action at} \Input{$f_\theta$ --- value and policy network}
\Output{$a \in \cA_i$ --- action for the current infostate}
Initialize $\mathbb{T}$, a hashmap with information per infostate.\\
\For{$i = 1, \ldots, n$}{
  $h = \textsc{SampleHistoryFromInfostate}(s, \pi_{-i})$.\\
  $\cT, r = \textsc{Simulation}(h, f_\theta, \pi_{-i}, \mathbb{T})$ \label{algline:mcts}\\
  $\textsc{UpdateSearchTree}(\cT, r)$.}
$\cN = \textsc{GetNode}(s, \mathbb{T})$\\
$\boldsymbol{p} = \textsc{NormalizedVisitCounts}(\cN)$\\
$a = \textsc{ChooseAction}(\cN)$\\
\Return{$(a, \boldsymbol{p})$};
\end{algorithm}

\begin{algorithm}[tb]
\caption{\textsc{Simulation} from Alg. \ref{alg:ismctsbr}.} 
\label{alg:simulation}
\SetKwInOut{Input}{input}\SetKwInOut{Output}{output}
\Input{$h$ --- root history for simulation}
\Input{$\pi_{-i}$ --- fixed policy for opponent}
\Input{$f_\theta$ --- policy \& value function} 
\Input{$\mathbb{T}$ --- hashmap with information per infostate} 
\Input{$f_\theta$ --- value and policy network}
\Output{$a \in \cA_i$ --- action for the current infostate}
Initialize $\cT$.\\
\While{$h$ is not terminal OR $h$ is not in $\mathbb{T}$} {
    $\cN = \textsc{GetNode}(h, \mathbb{T})$\\
    Add $\cN$ to $\cT$.\\
    \eIf{$i$ is to act} {$a$ = PUCB($\cN$).}
     {$a = \pi_{-i}(h)$}
    $h = ha$\\
}
\eIf{$h$ is terminal} {$r$ is the terminal value from $h$} {
    Add $h$ to $\mathbb{T}$.\\
    $r$ = value from $f_\theta(h)$
}
\Return{$(\cT, r)$};
\end{algorithm}

We call this algorithm \defword{ISMCTS-BR} and present it in Algorithm~\ref{alg:ismctsbr}.  $\textsc{GetNode}$ retrieves the node from the search tree corresponding to the information state $s$. We abuse notation and let $\textsc{GetNode}$ operate on histories $h$ as well, in which case $\textsc{GetNode}$ retrieves the node corresponding to the information state for the history from the current player's perspective.  $\textsc{SampleHistoryFromInfostate}$ samples a history $h$ (i.e. private information for the opponent) from the exact belief distribution induced by the opponent's policy. In imperfect information games with perfect recall, the belief $\Pr(h|s)$ is a function of the opponent's policy and can be obtained by applying Bayes' rule given the reach probabilities of each history (see~\cite[Section 3.2]{srinivasan2018actor}).
$\cT$ and $r$ are used to updated the search tree before proceeding to the next simulation. $\textsc{UpdateSearchTree}$ and $\textsc{ChooseAction}$ follow the standard MCTS algorithm~\cite{Silver18AlphaZero}; an IS-MCTS implementation can be found in \cite{LanctotEtAl2019OpenSpiel}. 

Running ISMCTS-BR requires having access to the opponent policy, as we query the opponent policy during Alg. \ref{alg:simulation} when the opponent would act. However, in practice, this is an easy restriction to relax, as replacing the exact opponent policy with a different policy weakens the quality of the resulting BR, but the result is still an approximate best response. For example, if an opponent policy is slow to query, we can train a function approximator to approximate the policy, and query the approximation during each simulation. If we only have access to a ``black box'' policy that returns an action, rather than a probability distribution, we can either train a function approximator or query the policy repeatedly; either would approximate the probability distribution. This generality is an advantage of the approximate best response framework: the proof is in the pudding, so to speak, as each ABR is a lower bound on how exploitable the given policy is. As such, one can choose from multiple different ABRs depending on the circumstance.

\section{Experimental Evaluation}
\label{sec:evaluation}

We use an experimental setup identical to \cite{Silver18AlphaZero} except where otherwise noted. We train for only 100k learning steps rather than 800k steps. We use a distributed actor/learner setup to train our neural network. Each experiment requires roughly 128 Cloud TPUv4 chips for the actors, and 4 TPUv2 chips for the learners. The TPU version choice was arbitrary and based on internal availability at the time we ran our experiments. The results we report, unless otherwise indicated, are the average results over the last five network updates in the run; as the network is updated every 500 minibatch steps, and each minibatch uses 2048 examples, this is roughly 5 million states of data. On iteration $t$, network parameters $\boldsymbol{\theta}_t$ are updated using a loss function that combines the mean-squared error between predicted expected reward $v_t$ and the Monte Carlo return for the episode $z_t$, the cross-entropy loss between the prior policy predicted by the network $\textbf{p}_t$ and the policy induced by the normalized visit counts during search $\boldsymbol{\pi}_t$, and $\ell_2$ regularization:
\begin{align}
(\mathbf{p}_t, v_t) &= f_{\theta_t}(s), \label{eq:vp}\\
l_t &= (z_t - v_t)^2 - \boldsymbol{\pi}^T_t \log \mathbf{p}_t + c \lVert \boldsymbol{\theta}_t \rVert_2^2 \\
\boldsymbol{\theta}_t &= \textsc{GradDescent}(\boldsymbol{\theta}_{t-1}, \alpha_t, l_t) 
\end{align}
Since the reward and prior are learned for information states, not histories, the network learns an average across histories, allowing the agent to act according to the expected value of information state $s$.

We present results in imperfect information games, large perfect information games, Scotland Yard (a classic pursuit-evasion game on a graph), and two large variants of Texas hold'em poker: heads-up limit (HUL) and heads-up no-limit (HUNL).
As these are widely-used domains, we provide descriptions in the appendix. 
Opponents range from basic and easy to exploit (e.g. uniform random) to agents that use search and are hard to exploit (e.g. AlphaZero).

\subsection{Validation}
\label{sec:validation}

We begin by validating ISMCTS-BR with policies \& domains that are simple enough to allow us to compute the exact exploitability.

\begin{table}[]
\centering
\begin{tabular}{lr@{\hskip1pt}l}
\toprule
Game                         & Unif. Rand. & \\ \midrule
Leduc Poker              & 98\%&        \\
Goofspiel4               & 97\%&        \\
Goofspiel5               & 95\%&        \\
Goofspiel6               & 97\%&        \\
HUL.                     & 90\%&\footnoteref{foot:hul-exploitability}        \\
SY - Glasses             & 99\%&        \\
SY - War Museum          & 96\%&        \\
SY - Queen Mary's Garden & 98\%&        \\
SY - Kennington Double   & 99\%&        \\
Full Scotland Yard       & 99\%&\footnoteref{foot:uniform-exploitability}          \\
Chess                    & 90\%&\footnoteref{foot:uniform-exploitability}        \\
Go                       & 100\%&\footnoteref{foot:uniform-exploitability}       \\
Connect 4                & 100\%&       \\ \bottomrule
\end{tabular}
\caption{ISMCTS-BR ANC (\%) vs UniformRandom.}
\label{tab:abr-uniform-random}
\end{table}
\footnote{\cite{Johanson2011accelerating-uniform}\label{foot:hul-exploitability}}
\paragraph{vs. Uniform Random.} We evaluate ISMCTS-BR versus uniform random in numerous games. Table~\ref{tab:abr-uniform-random} shows that ISMCTS-BR consistently performs well, achieving $\ge$95\% of achievable reward in most cases.\footnote{We assume that an exact BR to uniform wins 100\% in full Scotland Yard, Chess, and Go, as this value is intractable to compute.\label{foot:uniform-exploitability}}  GoofspielN refers to a variant of Goofspiel with N cards. SY refers to Scotland Yard. ``Glasses'', ``War Museum'', ``Queen Mary's Garden'' and ``Kennington Double'' refer to custom graphs of Scotland Yard that are small enough to compute exact NashConv (see the appendix for the specific graphs).

\begin{table}[]
\centering
\begin{tabular}{lrrrr} \toprule
            & Fold                      & Raise                      & $\frac{1}{2}$ Call $\frac{1}{2}$ Raise               & Call                     \\
            \midrule
Leduc Poker & 100\% & 99\% & 100\% & 99\%\\
HUL         & 100\% & 89\% & 94\%  & 97\%\\
HUNL(fcpa)  & 100\% & $\ge$77\%\footnoteref{foot:poker-exploitability} & $\ge$95\%\footnoteref{foot:poker-exploitability}  & 99\%\\
\bottomrule
\end{tabular}
\caption{ABR vs poker chumps, percentage of the exact BR value achieved. HUNL(fcpa) limits actions to fold, call, pot bet, and all-in.}
\label{tab:abr-poker-chumps}
\end{table}

\paragraph{vs. Poker Chumps.} Simple poker policies called ``chumps'' have been evaluated in prior work. These policies either always fold, always call, always raise, or mix between call and raise with equal probability.
Table~\ref{tab:abr-poker-chumps} presents the results as a percentage of the value achieved by an exact best response\footnote{Leduc poker is small making exact exploitability simple, while values for HUL come from~\cite[Table 1]{johanson2011accelerating}.  In HUNL, the prior LBR work~\cite{Lisy16LBR} argues that always call has an exploitability of approximately 50 bb/h (big blinds per hand).  We use this value for all chumps except always fold, as the same argument holds as an upper-bound on exact exploitability, likely causing an underestimate of ISMCTS-BR's effectiveness.\label{foot:poker-exploitability}}.  In most cases, the reward achieved by ISMCTS-BR quickly converges to the same value as the exact best response. Practitioners should note that this evaluation was particularly helpful in developing ISMCTS-BR as it enabled testing on a range of behaviour at different game scales using exact exploitability values that were generally known.

We can compare LBR with ISMCTS-BR and the exact BR against chump policies in poker. Table~\ref{table:chumps} shows that ISMCTS-BR is able to improve upon LBR in HUL, except for the case of always call, where LBR's value function exactly matches the opponent's behaviour.  LBR is notably weaker against uniform random.  Results in HUNL echo this, with ISMCTS-BR performing similarly against always call (49.7 bb/h instead of LBR's 49.0), but performing better than the best LBR result reported against the call/raise chump in~\cite{Lisy16LBR}: 47.9 bb/h instead of LBR's 24.4. This is a shortcoming of LBR: the specific choice of heuristic introduces bias, and it can fail to optimally exploit weak agents.\footnote{We ran the LBR results using an implementation of \cite{Lisy16LBR}. Results are the mean over 200 000 games, and all have a standard error less than 30 mbb/h.}

\begin{table}
\centering
\begin{tabular}{lrrr}
\toprule
Policy        & ANC      & LBR   & NashConv \\
\midrule
Fold          & 1500     & 1500  & 1500 \\
Call          & 2270     & 2310  & 2330 \\    
Random        & 7960     & 5690  & 8800 \\
\bottomrule
\end{tabular}
\caption{Comparison of (approximate) NashConv values for HUL chumps. Units are in mbb/h. ANC is ISMCTS-BR's ApproximateNashConv.}
\label{table:chumps}
\end{table}

\begin{table}[]
\centering
\begin{tabular}{lr}
\toprule
                         & CFR+10\\
                         \midrule
Goofspiel4               & 99.9\%\\
Goofspiel5               & 95.7\%\\
Goofspiel6               & 99.5\%\\
SY - Glasses             & 94.9\%\\
SY - War Museum          & 99.9\%\\
SY - Queen Mary's Garden & 99.6\%\\
SY - Kennington Double   & 97.6\%\\
\bottomrule
\end{tabular}
\caption{Approx. NashConv (\% of exact NashConv) for ISMCTS-BR vs CFR+ with 10 iters.}
\label{tab:abr-cfr}
\end{table}

\paragraph{vs. Nash Approximations.}  One limitation of the previous opponents is that they behave the same regardless of the game's state. To address this, we examine ISMCTS-BR in several small imperfect information games against Nash approximations generated with 10 iterations of CFR+~\cite{Tammelin15CFRPlus}, ensuring the policy remains exploitable. Results in Table~\ref{tab:abr-cfr} show that ISMCTS-BR can exploit these more sophisticated policies. Against CFR+ with more than 10 iterations, ISMCTS-BR also yields exploitability close to the exact value. However, as the exploitability approaches zero with more iterations, small absolute variations cause large percentage variations.

\begin{figure}[!ht]
\centering
\begin{tabular}{@{}c@{}}
    \includegraphics[width=.48\textwidth]{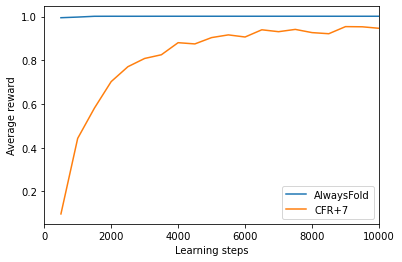} 
  \end{tabular}
  \caption{ISMCTS-BR training curves in Leduc poker vs. always fold and CFR+ with 7 iterations.}
  \label{fig:cfr-alwaysfold}
\end{figure}

Nash approximations also provide an opportunity to demonstrate the richness of evaluation enabled by learned best responses.  We trained ISMCTS-BR in Leduc poker against CFR+ with 7 iterations and always fold, which have exact exploitabilities of 1.01 and 1, respectively. If practitioners only consider the exact exploitability, or the final exploitability produced by ABR, these policies would appear similar. However, Figure~\ref{fig:cfr-alwaysfold} shows that CFR+7 is more difficult to exploit --- reflecting the strategy's relative complexity.

Finally, we ran ISMCTS-BR vs Slumbot \cite{jackson2013slumbot}, a strong agent in heads-up no-limit Texas hold'em. We evaluate a best effort reproduction of Slumbot 2016\footnote{Slumbot's author, Eric Jackson, provided data files for Slumbot 2016 that were used with up-to-date code and a new configuration he believed best matched Slumbot 2016.} using ISMCTS-BR and compare performance to prior LBR results for the original Slumbot 2016~\cite{Lisy16LBR}. ISMCTS-BR managed to exploit Slumbot by 1259 mbb/h when ABR acted using the FCPA betting abstraction.  This outperforms the prior LBR results for responses that were not forced to check/call through the first two rounds, which won 522 and 763 mbb/h using the fold/call and 56 bets action abstractions, respectively.  By constraining LBR's actions per round --- limiting it to check/call until the third round --- LBR was able to win 4020 and 3763 mbb/h using the fold/call/pot/all-in and 56 bets action abstractions.  While ISMCTS-BR uses the fcpa abstraction, we have not restricted it per round. Prior LBR results do not evaluate the fcpa abstraction without restrictions, but the contrast in performance when using the 56 bets abstraction is considerable, both against Slumbot and the other agents that were evaluated, suggesting that this domain-specific hand-tuning was important.  The generality of ISMCTS-BR allows it to find a substantial degree of exploitability, even in high-quality agents, across disparate domains and despite a lack of such domain knowledge.

\subsection{Approximate Exploitability in Go}
\label{sec:AZ-go}



We use ISMCTS-BR to evaluate two agents trained following \cite{Silver18AlphaZero}: one as described in the paper (AlphaZero MSE), and another using the 25-step TD error to train the value function (AlphaZero TD25).\footnote{We discuss the strength of these agents in the appendix.} During the ISMCTS-BR search, we use the AlphaZero policy head as the opponent policy, rather than the full MCTS algorithm. When AlphaZero is selecting actions during play, it acts as normal, and uses 800 simulations.

\begin{figure}[!ht]
\centering
\begin{tabular}{@{}c@{}}
    \includegraphics[width=.45\textwidth]{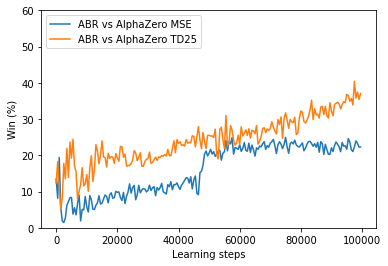} 
  \end{tabular}
  \caption{ISMCTS-BR results vs AlphaZero agents in Go.}
  \label{fig:abr-az}
\end{figure}

In head-to-head play, AlphaZero MSE wins 23.25\% of games vs AlphaZero TD25, indicating a relative Elo of\\ -207.  AlphaZero MSE was found to be less exploitable, with ISMCTS-BR at the end of training winning roughly 20\% of games instead of the 40\% against AlphaZero TD25 (Fig~\ref{fig:abr-az}).

\begin{figure}[!ht]
\centering
\begin{tabular}{@{}c@{}}
    \includegraphics[width=.45\textwidth]{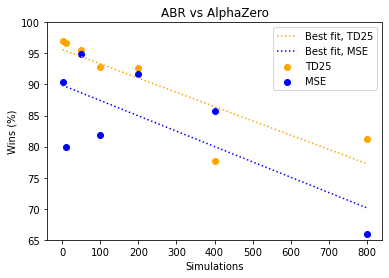} 
  \end{tabular}
  \caption{ISMCTS-BR simulation curriculum results in Go against AlphaZero agents.}
  \label{fig:abr-az-expl-curric}
\end{figure}

As AlphaZero is a MCTS agent with a variable number of simulations, we can create a curriculum --- training ISMCTS-BR against AlphaZero with 1 simulation, then warm-starting the network and train it against AlphaZero with 10 simulations, then 50, 100, 200, 400, and 800. (Fig. ~\ref{fig:abr-az-expl-curric}). The curriculum improved the exploitability estimate significantly, returning results around 90\%.

One common way to determine whether an algorithmic change is an improvement is to compare head-to-head performance play with and without the change. If that was done to choose between the TD25 and MSE variants, one would neglect the agent's worst-case performance, potentially deploying an agent that is vulnerable to failure in the wild.  Head-to-head performance and exploitability are complementary, and practitioners much decide how practical worst-case scenarios are.  Perhaps limited access to the agent makes it unrealistic for adversaries to probe its behaviour, allowing for more focus on the in-practice head-to-head performance.

\section{Conclusion}

We introduce ISMCTS-BR, an algorithm for learning an approximate best response that generalizes the poker specific local best response method, and validate it across a range of domains and agents.  In addition to a lower-bound on the true exploitability, a learned best response provides both an assessment of how difficult it is to discover an agent’s flaws, and a mechanism to facilitate debugging by identifying game trajectories where the agent is weak.
We believe this technique will support evaluation in large games and drive algorithmic improvements for robust agents, as prior work spurred computer poker research.

Although ISMCTS-BR is limited to tractable belief spaces, this could potentially be relaxed by using a sample model to generate actions for the opponent during search or replacing search with a stochastic planning algorithm such as \cite{ozair2021vector} as the best responder.

\section*{Acknowledgements}

We would like to thank Eric Jackson for helping us evaluate against Slumbot, Kevin Waugh for repeatedly finding \& fixing bugs in our code, and Josh Davidson for providing the Scotland Yard images.

\bibliographystyle{named}
\bibliography{paper}

\include{appendix}

\end{document}

%% file: appendix.tex
\begin{appendix}

\section{Game Domain Descriptions}

Chess, Connect Four, and Go are classic challenge domains used in AI. They are all implemented in the standard way, as available in OpenSpiel~\cite{LanctotEtAl2019OpenSpiel}. Go uses the standard rules as used by AlphaZero~\cite{Silver18AlphaZero}.

{\bf Leduc Poker} is a simplified poker game with a six-card deck containing three ranks and two suits. Players must each ante one chip to play. Each player receives a single private card, and there are two betting rounds with legal actions: fold, call, and raise. There is a limit of two raised per round. Raises are set to 2 chips in the first round and 4 chips in the second round. The utility is the difference in a player's chip after the game ends and and before the game started.

{\bf Goofspiel}N is a bidding card game where players start with $N$ bid cards numbered $1 \cdots N$. There is also a deck of point cards, face down, numbered $1 \cdots N$. Every turn, the top card of the point card deck is revealed, and each player chooses a single bid card to play simultaneously. The winner of the turn is the player with the highest bid, and that player receives the face-up point card. At the end of the game, the player with the highest number of points wins.
We use a common imperfect information variant where the bid cards are never revealed, only whether the turn was won or lost by each player. The exact OpenSpiel game string used is:
{\tt turn\_based\_simultaneous\_game(
game=goofspiel(imp\_info=true,num\_cards=N,
\\players=2,points\_order=descending,
returns\_type=win\_loss))}

{\bf Texas hold'em poker} is a popular poker variant played with a 52-card deck. At the start of each round, players receive two private cards and proceed to several betting rounds preceeding and following the revealing of public community cards: flop (three cards), turn (one card), and river (one card). 
Heads-up Limit (HUL) Texas hold'em poker restricts the bets and amounts to specific sizes in a similarly to Leduc poker.
No-limit hold'em (HUNL) removes the fixed bets restriction. However in this paper, we use a restricted variant that allows fold (f), call (c), pot-sized bet (p), and all-in (a) which we abbreviate HUNL(fcpa).
The game rules are based on the standard Annual Computer Poker Competition rules, and their OpenSpiel game strings are:
\begin{itemize}
    \item {\tt universal\_poker(betting=limit,numPlayers=2,
    numRounds=4,blind=10 5,\\firstPlayer=2 1,numSuits=4,numRanks=13,
    numHoleCards=2,\\numBoardCards=0 3 1 1,raiseSize=10 10 20 20,
    \\maxRaises=3 4 4 4,bettingAbstraction=full)}
    \item {\tt universal\_poker(betting=nolimit,numPlayers=2,
    numRounds=4,blind=100 50,\\firstPlayer=2 1 1 1,
    numSuits=4,numRanks=13,numHoleCards=2,\\numBoardCards=0 3 1 1,
    stack=20000 20000,bettingAbstraction=fcpa)}
\end{itemize}

\subsection{Scotland Yard}

Scotland Yard is a pursuit-evasion game played on a graph. It is typically played with 6 players--- 5 players playing as "detectives" who chase a fugitive, "Mr. X" throughout a stylized version of London. The Player of Games paper \cite{schmid2021player} has a description of the game. 

\begin{figure*}[h!]
\centering
\begin{tabular}{cc}
\includegraphics[scale=0.25]{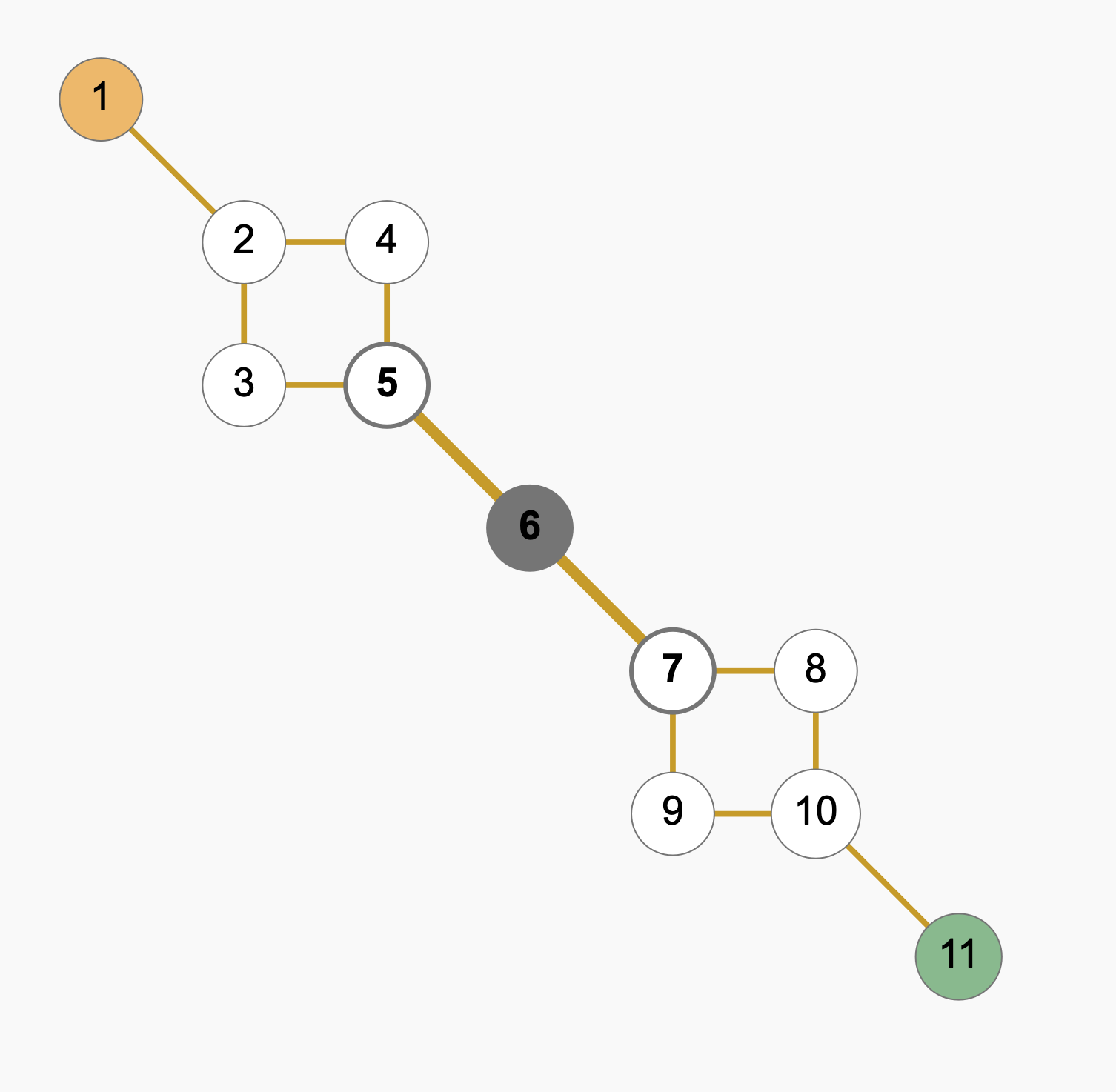} &
\includegraphics[scale=0.38]{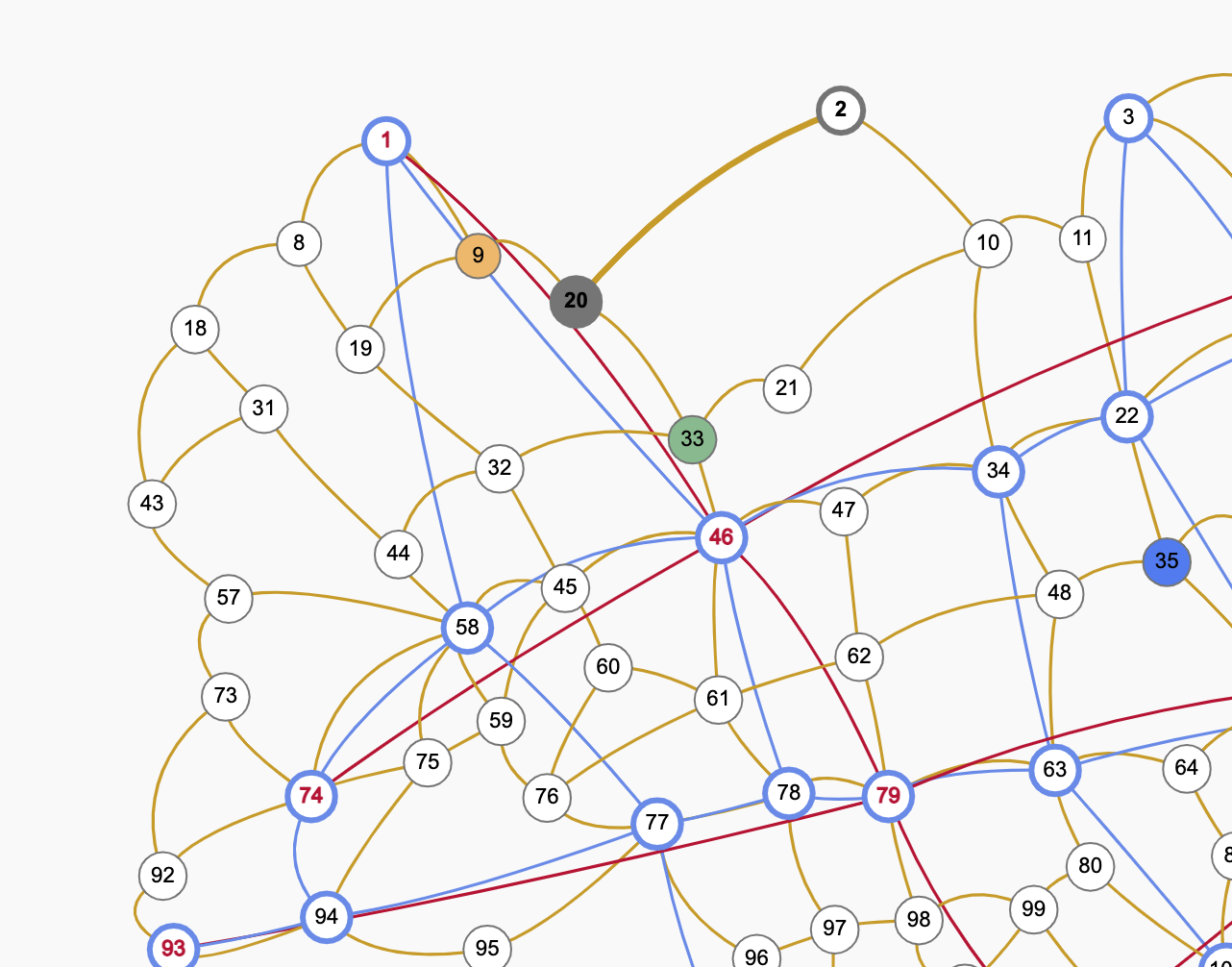} \\
Glasses & Queen Mary's Garden \\
\includegraphics[scale=0.35]{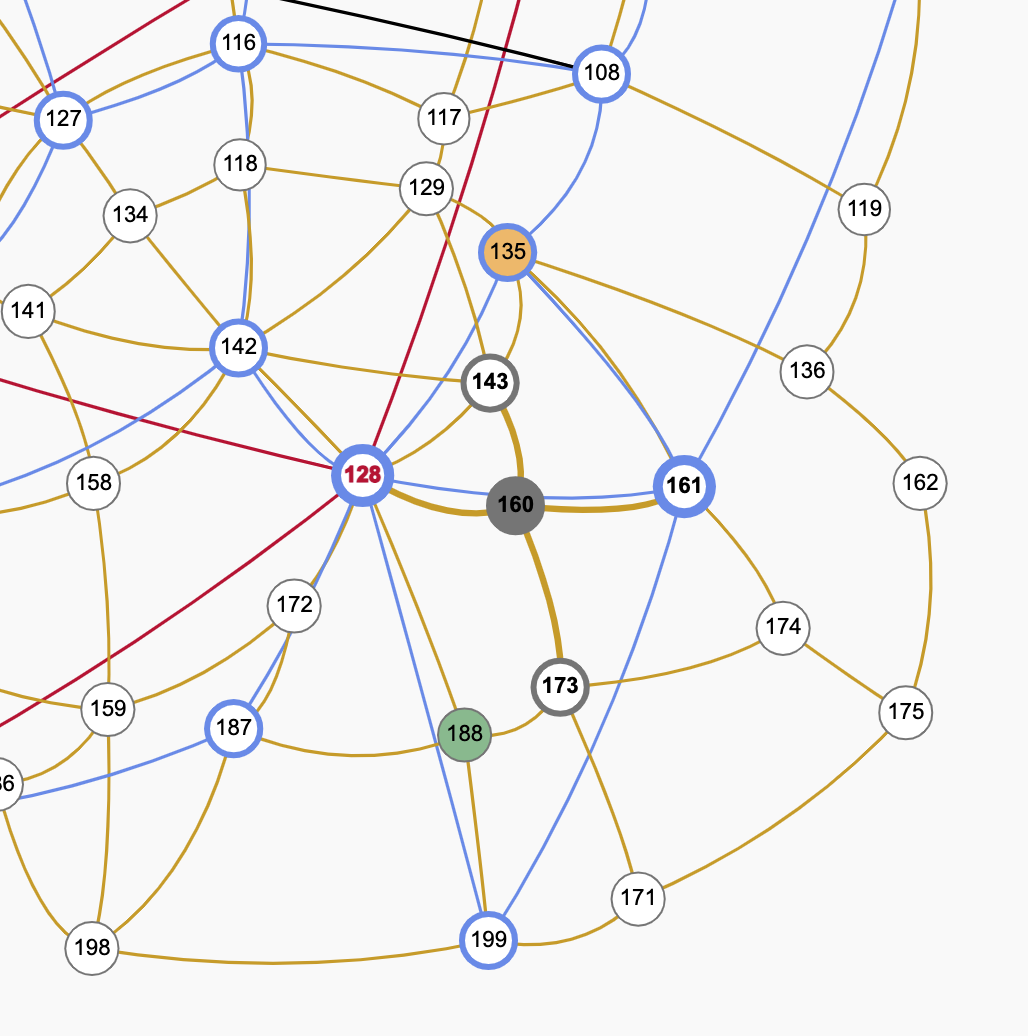} &
\includegraphics[scale=0.38]{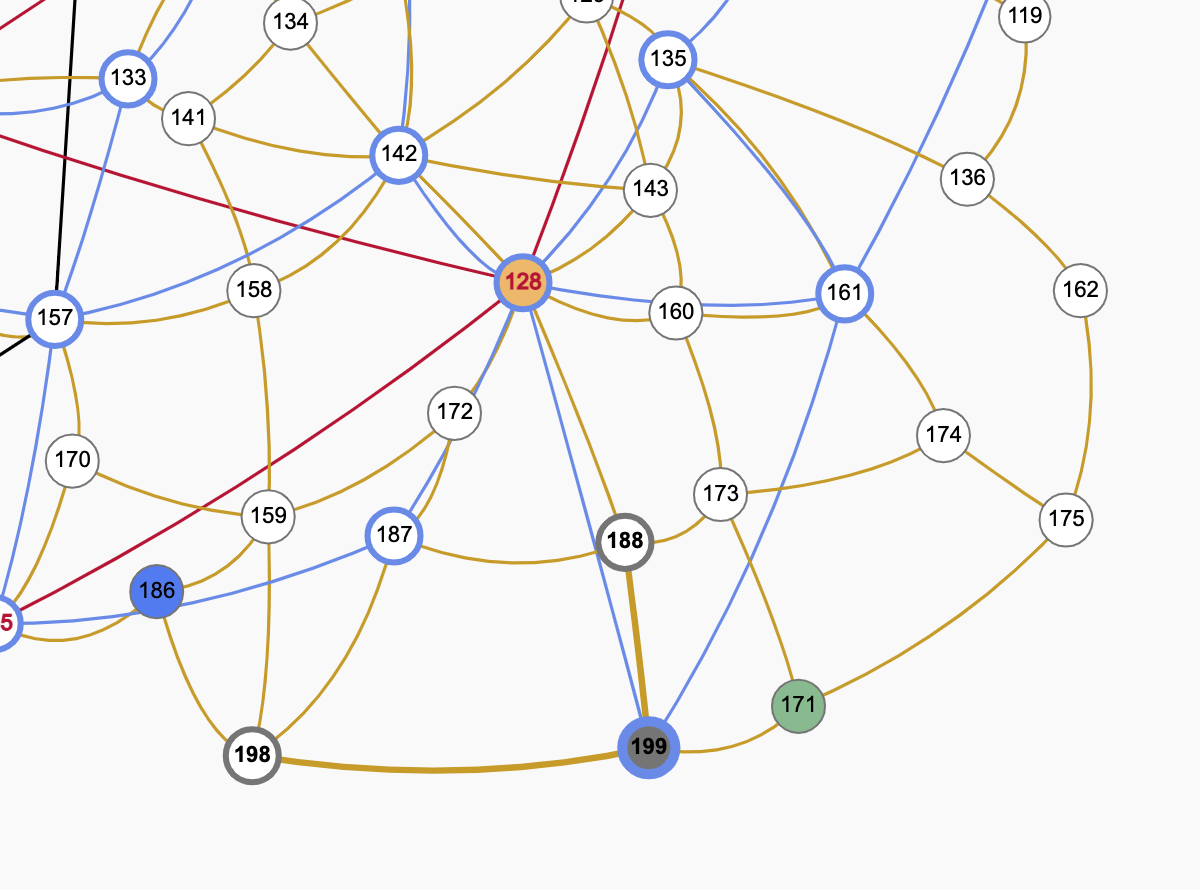} \\
War Museum & Kennington Double \\
\end{tabular}
 \caption{Scotland Yard custom maps. The darkest-shaded location is Mr. X.'s location whereas the lighter-colored locations represent the detectives starting locations. Edges colors represent different transport types (taxi, bus, underground). Color meaning is in Table~\ref{tab:sy-colors}.}
    \label{fig:sy-maps}
\end{figure*}

We introduce several smaller variants of Scotland Yard (see Figure~\ref{fig:sy-maps} below for a visualization):

\begin{itemize}
    \item Glasses, a custom graph that looks like a pair of glasses.
    \item War Museum, a subgraph of the full game centered around the War Museum.
    \item Queen Mary's Garden, a subgraph of the full game centered around Queen Mary's Garden.
    \item Kennington Double, a subgraph of the full game set in Kennington with a double move available to Mr. X.
\end{itemize}

In each, we create new graphs that the players must traverse. In each, some number of detectives have a limited number of moves to find Mr. X. In Table \ref{tab:sy-subgame-mrx} and \ref{tab:sy-subgame-det} we include details about each subgame. Unlike the full game, in the small variants, Mr. X is restricted to only use taxi tickets. The detectives have additional tickets. The full game also includes boats, which are only usable by Mr. X. The shaded nodes are those with a player on them; the grey shaded node is Mr. X, and the other shaded nodes are the detectives. For instance, in Glasses, Mr. X starts on node 6, while detectives start on nodes 1 and 11.

\begin{table*}[t!]
\centering
\begin{tabular}{lrrr}
\toprule
Variant             & \multicolumn{1}{l}{MrX Taxi tickets} & \multicolumn{1}{l}{MrX double moves} & \multicolumn{1}{l}{Number of rounds} \\ \midrule
Glasses             & 5                                    & 0                                    & 5                                    \\
Queen Mary's Garden & 3                                    & 1                                    & 3                                    \\
War Museum          & 1                                    & 0                                    & 1                                    \\
Kennington Double   & 2                                    & 1                                    & 2                                    \\ \bottomrule
\end{tabular}
\caption{Scotland Yard subgame description for Mr. X.}
\label{tab:sy-subgame-mrx}
\end{table*}

\begin{table*}[t!]
\centering
\begin{tabular}{lrrrr}
\toprule
Variant             & \# detectives & Taxi tickets & Bus  tickets & Underground  tickets \\ \midrule
Glasses             & 2                                        & 5                                          & 0                                         & 0                                                 \\
Queen Mary's Garden & 3                                        & 3                                          & 0                                         & 0                                                 \\
War Museum          & 2                                        & 1                                          & 1                                         & 1                                                 \\
Kennington Double   & 3                                        & 2                                          & 2                                         & 2                                                 \\ \bottomrule
\end{tabular}\caption{Scotland Yard subgame description for the detectives.}
\label{tab:sy-subgame-det}
\end{table*}

\begin{table}[]
\centering
\begin{tabular}{lr}
\toprule
Ticket type & color \\ \midrule
Taxi & yellow\\
Bus & blue \\
Underground & red\\
Boat & black\\
\bottomrule
\end{tabular}\caption{Scotland Yard edge colors.}
\label{tab:sy-colors}
\end{table}

\newpage

\

\section{AlphaZero agents}

The two AlphaZero agents we train ISMCTS-BR to exploit were trained with the current version of the AlphaZero/MuZero codebase. The code is largely identical to that used to train the agents from \cite{Silver18AlphaZero}, but the codebase has been actively developed since 2018, and as such, there have been some changes. For our experimental setup, the sole major deviation from \cite{Silver18AlphaZero} is in the hardware. As noted in \cite{Silver18AlphaZero}:

\begin{quote}
During training only, 5000 first-generation tensor processing units (TPUs) were used to generate self-play games, and 16 second-generation TPUs were used to train the neural networks.
\end{quote}

The two AlphaZero agents we trained used 4000 third-generation TPUs to generate self-play games, and 16 second-generation TPUs were used to train the neural networks. The reason for this is that the first-generation TPUs were unavailable for us to use at the time of training.\\ 

The agent we refer to as "AlphaZero MSE" is the same agent as is referred to as "AlphaZero" in \cite{schmid2021player}, which includes additional strength comparisons; we include a subset here in Table \ref{tab:go-elo-comparisons}. "PoG" refers to the Player of Games agent introduced in \cite{schmid2021player}. Pachi was introduced in \cite{baudivs2011pachi}; we use version 11.99. We also evaluate against level 10 GnuGo, version 3.8.

In Table \ref{tab:go-elo-comparisons}, we estimate the strength of AlphaZero TD25 here by transitivity in relation to AlphaZero MSE; as AlphaZero TD25 beats AlphaZero MSE 76.75\% of the time, we calculate an Elo value for it as 2099 + 207 = 2306.

\begin{table}[]
\centering
\begin{tabular}{lr}
\toprule
Agent & Relative Elo\\
\midrule
AlphaZero TD25 (sims=800, timesteps=800k) & 2306 \\
AlphaZero MSE (sims=800, timesteps=800k)  & 2099 \\
PoG(s=800, c=1) & 1426  \\
Pachi, 100k iterations     & 869   \\
Pachi 10k iterations     & 231 \\
GnuGo Level 10 & 0    \\
\bottomrule
\end{tabular}
\caption{Relative strength comparisons for the Go agents that we trained. All results other than AlphaZero TD25 come from Table 4 of \protect\cite{schmid2021player}, as we use the same agents.}
\label{tab:go-elo-comparisons}
\end{table}

\end{appendix}

%% file: paper.bbl
\begin{thebibliography}{}

\bibitem[\protect\citeauthoryear{Allis}{1994}]{Allis94thesis}
V.~L. Allis.
\newblock {\em Searching for solutions in games and artificial intelligence}.
\newblock PhD thesis, University of Limburg, 1994.

\bibitem[\protect\citeauthoryear{Amodei \bgroup \em et al.\egroup
  }{2016}]{AmodeiOSCSM16Safety}
D.~Amodei, C.~Olah, J.~Steinhardt, P.F. Christiano, J.~Schulman, and
  D.~Man{\'{e}}.
\newblock Concrete problems in {AI} safety.
\newblock {\em CoRR}, abs/1606.06565, 2016.

\bibitem[\protect\citeauthoryear{Athalye \bgroup \em et al.\egroup
  }{2018}]{AthalyeEIK18}
A.~Athalye, L.~Engstrom, A.~Ilyas, and K.~Kwok.
\newblock Synthesizing robust adversarial examples.
\newblock In {\em Proceedings of the 35th International Conference on Machine
  Learning (ICML)}, pages 284--293. {PMLR}, 2018.

\bibitem[\protect\citeauthoryear{Balduzzi \bgroup \em et al.\egroup
  }{2018}]{Balduzzi18}
D.~Balduzzi, K.~Tuyls, J.~P{\'{e}}rolat, and T.~Graepel.
\newblock Re-evaluating evaluation.
\newblock {\em CoRR}, abs/1806.02643, 2018.

\bibitem[\protect\citeauthoryear{Bard}{2016}]{bard2016online}
N.~DC Bard.
\newblock {\em Online Agent Modelling in Human-Scale Problems}.
\newblock PhD thesis, U. Alberta, 2016.

\bibitem[\protect\citeauthoryear{Baudi{\v{s}} and
  Gailly}{2011}]{baudivs2011pachi}
Petr Baudi{\v{s}} and Jean-loup Gailly.
\newblock Pachi: State of the art open source go program.
\newblock In {\em Advances in computer games}, pages 24--38. Springer, 2011.

\bibitem[\protect\citeauthoryear{{Bellemare} \bgroup \em et al.\egroup
  }{2013}]{bellemare13arcade}
M.~G. {Bellemare}, Y.~{Naddaf}, J.~{Veness}, and M.~{Bowling}.
\newblock The arcade learning environment: An evaluation platform for general
  agents.
\newblock {\em Journal of Artificial Intelligence Research}, 47:253--279, jun
  2013.

\bibitem[\protect\citeauthoryear{Brown and Sandholm}{2017}]{Brown17Libratus}
N.~Brown and T.~Sandholm.
\newblock Superhuman {AI} for heads-up no-limit poker: {L}ibratus beats top
  professionals.
\newblock {\em Science}, 360(6385), 2017.

\bibitem[\protect\citeauthoryear{Brown \bgroup \em et al.\egroup
  }{2018}]{brown2018depth}
N.~Brown, T.~Sandholm, and B.~Amos.
\newblock Depth-limited solving for imperfect-information games.
\newblock {\em Advances in NeurIPS}, pages 7663--7674, 2018.

\bibitem[\protect\citeauthoryear{Campbell \bgroup \em et al.\egroup
  }{2002}]{campbell2002deep}
M.~Campbell, A.~J.~Hoane Jr, and F.~Hsu.
\newblock Deep blue.
\newblock {\em Artif. intell.}, 134(1-2):57--83, 2002.

\bibitem[\protect\citeauthoryear{Cowling \bgroup \em et al.\egroup
  }{2010}]{Cowling12ISMCTS}
P.I. Cowling, E.J. Powley, and D.~Whitehouse.
\newblock Information set {M}onte {C}arlo tree search.
\newblock {\em IEEE Transactions on Computational Intelligence and AI in
  Games}, 4(2):120--143, 2010.

\bibitem[\protect\citeauthoryear{Gleave \bgroup \em et al.\egroup
  }{2020}]{GleaveDWKLR20}
A.~Gleave, M.~Dennis, C.~Wild, N.~Kant, S.~Levine, and S.~Russell.
\newblock Adversarial policies: Attacking deep reinforcement learning.
\newblock In {\em 8th Int'l Conference on Learning Representations, {ICLR}},
  2020.

\bibitem[\protect\citeauthoryear{Greenwald \bgroup \em et al.\egroup
  }{2017}]{greenwald2017solving}
A.~Greenwald, J.~Li, and E.~Sodomka.
\newblock Solving for best responses and equilibria in extensive-form games
  with reinforcement learning methods.
\newblock In {\em Rohit Parikh on Logic, Language and Society}, pages 185--226.
  Springer, 2017.

\bibitem[\protect\citeauthoryear{Gruslys \bgroup \em et al.\egroup
  }{2020}]{armac}
A.~Gruslys, M.~Lanctot, R.~Munos, F.~Timbers, M.~Schmid, and et~al.
\newblock The advantage regret-matching actor-critic.
\newblock {\em ArXiv}, 2020.

\bibitem[\protect\citeauthoryear{Jackson}{2013}]{jackson2013slumbot}
E.G. Jackson.
\newblock Slumbot {NL}: Solving large games with counterfactual regret
  minimization using sampling and distributed processing.
\newblock In {\em Workshops of the Twenty-Seventh AAAI Conference}, 2013.

\bibitem[\protect\citeauthoryear{Johanson \bgroup \em et al.\egroup
  }{2011}]{johanson2011accelerating}
M.~Johanson, K.~Waugh, M.~Bowling, and Martin M.~Zinkevich.
\newblock Accelerating best response calculation in large extensive games.
\newblock In {\em IJCAI}, 2011.

\bibitem[\protect\citeauthoryear{Johanson}{2011}]{Johanson2011accelerating-uniform}
M.~Johanson.
\newblock Accelerating best response calculation in large extensive games.
\newblock
  http://johanson.ca/publications/poker/2011-ijcai-abr/2011-ijcai-abr.html,
  2011.

\bibitem[\protect\citeauthoryear{Johanson}{2013}]{johanson2013measuring}
M.~Johanson.
\newblock Measuring the size of large no-limit poker games.
\newblock {\em arXiv abs/1302.7008}, 2013.

\bibitem[\protect\citeauthoryear{Lanctot \bgroup \em et al.\egroup
  }{2019}]{LanctotEtAl2019OpenSpiel}
Marc Lanctot, Edward Lockhart, Jean-Baptiste Lespiau, Vinicius Zambaldi, and
  et~al.
\newblock {OpenSpiel}: A framework for reinforcement learning in games.
\newblock {\em CoRR}, abs/1908.09453, 2019.
\newblock \url{http://arxiv.org/abs/1908.09453}.

\bibitem[\protect\citeauthoryear{Lis{\'{y}} and Bowling}{2016}]{Lisy16LBR}
Viliam Lis{\'{y}} and Michael~H. Bowling.
\newblock Eqilibrium approximation quality of current no-limit poker bots.
\newblock {\em CoRR}, abs/1612.07547, 2016.

\bibitem[\protect\citeauthoryear{Lis\'{y} \bgroup \em et al.\egroup
  }{2015}]{Lisy15Online}
V.~Lis\'{y}, M.~Lanctot, and M.~Bowling.
\newblock Online {M}onte {C}arlo counterfactual regret minimization for search
  in imperfect information games.
\newblock In {\em Proceedings of the Int'l Conf. on Autonomous Agents and
  Multi-Agent Systems ({AAMAS})}, pages 27--36, 2015.

\bibitem[\protect\citeauthoryear{Mnih \bgroup \em et al.\egroup
  }{2015}]{mnih2015human}
V.~Mnih, K.~Kavukcuoglu, D.~Silver, A.~Rusu, and et~al.
\newblock Human-level control through deep reinforcement learning.
\newblock {\em nature}, 518(7540):529--533, 2015.

\bibitem[\protect\citeauthoryear{Morav{\v c}{\'\i}k \bgroup \em et al.\egroup
  }{2017}]{Moravcik17DeepStack}
M.~Morav{\v c}{\'\i}k, M.~Schmid, N.~Burch, and et~al.
\newblock Deepstack: Expert-level artificial intelligence in heads-up no-limit
  poker.
\newblock {\em Science}, 358(6362), 2017.

\bibitem[\protect\citeauthoryear{Omidshafiei \bgroup \em et al.\egroup
  }{2019}]{Omidshafiei19}
S.~Omidshafiei, C.~Papadimitriou, Piliouras G, K.~Tuyls, and et~al.
\newblock $\alpha$-rank: Multi-agent evaluation by evolution.
\newblock {\em Scientific Reports}, 9(1):9937, 2019.

\bibitem[\protect\citeauthoryear{Ozair \bgroup \em et al.\egroup
  }{2021}]{ozair2021vector}
S.~Ozair, Y.~Li, A.~Razavi, I.~Antonoglou, A.~van~den Oord, and O.~Vinyals.
\newblock Vector quantized models for planning.
\newblock {\em arXiv preprint arXiv:2106.04615}, 2021.

\bibitem[\protect\citeauthoryear{Quionero-Candela \bgroup \em et al.\egroup
  }{2009}]{Quionero-Candela09}
J.~Quionero-Candela, M.~Sugiyama, A.~Schwaighofer, and N.D. Lawrence.
\newblock {\em Dataset Shift in Machine Learning}.
\newblock The MIT Press, 2009.

\bibitem[\protect\citeauthoryear{Schaeffer \bgroup \em et al.\egroup
  }{1996}]{schaeffer1996chinook}
J.~Schaeffer, R.~Lake, P.~Lu, and M.~Bryant.
\newblock Chinook the world man-machine checkers champion.
\newblock {\em AI Magazine}, 17(1):21--21, 1996.

\bibitem[\protect\citeauthoryear{Schmid \bgroup \em et al.\egroup
  }{2021}]{schmid2021player}
Martin Schmid, Matej Moravcik, Neil Burch, Rudolf Kadlec, Josh Davidson, Kevin
  Waugh, Nolan Bard, Finbarr Timbers, Marc Lanctot, Zach Holland, et~al.
\newblock Player of games.
\newblock {\em arXiv preprint arXiv:2112.03178}, 2021.

\bibitem[\protect\citeauthoryear{Silver \bgroup \em et al.\egroup
  }{2018}]{Silver18AlphaZero}
David Silver, Thomas Hubert, and Julian~Schrittwieser et~al.
\newblock A general reinforcement learning algorithm that masters chess, shogi,
  and {G}o through self-play.
\newblock {\em Science}, 632(6419):1140--1144, 2018.

\bibitem[\protect\citeauthoryear{Srinivasan \bgroup \em et al.\egroup
  }{2018}]{srinivasan2018actor}
S.~Srinivasan, M.~Lanctot, V.~Zambaldi, and et~al.
\newblock Actor-critic policy optimization in partially observable multiagent
  environments.
\newblock In {\em Advances in NeurIPS}, pages 3422--3435, 2018.

\bibitem[\protect\citeauthoryear{Tammelin \bgroup \em et al.\egroup
  }{2015}]{Tammelin15CFRPlus}
O.~Tammelin, N.~Burch, M.~Johanson, and M.~Bowling.
\newblock Solving heads-up limit texas hold'em.
\newblock In {\em IJCAI}, 2015.

\bibitem[\protect\citeauthoryear{Tuyls \bgroup \em et al.\egroup
  }{2018}]{Tuyls18Generalized}
K.~Tuyls, J.~Perolat, M.~Lanctot, J.Z. Leibo, and T.~Graepel.
\newblock A generalized method for empirical game theoretic analysis.
\newblock In {\em Proceedings of the Int'l Conf. on Auto. Agents and Multiagent
  Systems (AAMAS)}, 2018.

\bibitem[\protect\citeauthoryear{Witty \bgroup \em et al.\egroup
  }{2018}]{WittyLTALJ18}
S.~Witty, J.K. Lee, E.~Tosch, A.~Atrey, M.L. Littman, and D.D. Jensen.
\newblock Measuring and characterizing generalization in deep reinforcement
  learning.
\newblock {\em CoRR}, abs/1812.02868, 2018.

\bibitem[\protect\citeauthoryear{Zinkevich \bgroup \em et al.\egroup
  }{2008}]{08nips-cfr}
M.~Zinkevich, M.~Johanson, M.~Bowling, and C.~Piccione.
\newblock Regret minimization in games with incomplete information.
\newblock In {\em Adv. in Neural Information Processing Systems 20}, pages
  905--912, 2008.

\end{thebibliography}
